\pdfoutput=1

\documentclass[11pt]{article}

\usepackage[preprint]{acl}

\usepackage{times}
\usepackage{latexsym}
\usepackage{booktabs}
\usepackage{amsmath}
\usepackage{amsfonts}

\usepackage[T1]{fontenc}

\usepackage[utf8]{inputenc}

\usepackage{microtype}

\usepackage{inconsolata}

\usepackage{graphicx}

\usepackage{tcolorbox}
\tcbuselibrary{skins}
\definecolor{lightgreen}{RGB}{225, 239, 217} 
\usepackage{etoolbox}
\AtBeginEnvironment{tcolorbox}{\small}

\usepackage{multirow}

\usepackage{graphics}
\usepackage{subcaption}

\title{CoT-Valve: Length-Compressible Chain-of-Thought Tuning}

 \author{Xinyin Ma\thanks{Equal contribution}, ~Guangnian Wan\footnotemark[1], ~Runpeng Yu, ~Gongfan Fang, ~Xinchao Wang\thanks{Corresponding Author} \\
         National University of Singapore\\
         \{maxinyin, guangnian\}@u.nus.edu, xinchao@nus.edu.sg}

\newcommand{\methodname}{CoT-Valve}

\begin{document}
\maketitle

\begin{abstract}
Chain-of-Thought significantly enhances a model's reasoning capability, but it also comes with a considerable increase in inference costs due to long chains. With the observation that the reasoning path can be easily compressed under easy tasks but struggle on hard tasks, we explore the feasibility of elastically controlling the length of reasoning paths with only one model, thereby reducing the inference overhead of reasoning models dynamically based on task difficulty. We introduce a new tuning and inference strategy named CoT-Valve, designed to allow models to generate reasoning chains of varying lengths. To achieve this, we propose to identify a direction in the parameter space that, when manipulated, can effectively control the length of generated CoT. Moreover, we show that this property is valuable for compressing the reasoning chain. We construct datasets with chains from long to short for the same questions and explore two enhanced strategies for CoT-Valve: (1) a precise length-compressible CoT tuning method, and (2) a progressive chain length compression approach. Our experiments show that \methodname\ successfully enables controllability and compressibility of the chain and shows better performance than the prompt-based control. We applied this method to QwQ-32B-Preview, reducing reasoning chains on GSM8K from 741 to 225 tokens with a minor performance drop (95.07\% to 94.92\%) and on AIME from 6827 to 4629 tokens, with only one additional incorrect answer.
\end{abstract}

\section{Introduction}

\begin{figure*}[t]
    \centering
    \includegraphics[width=\linewidth]{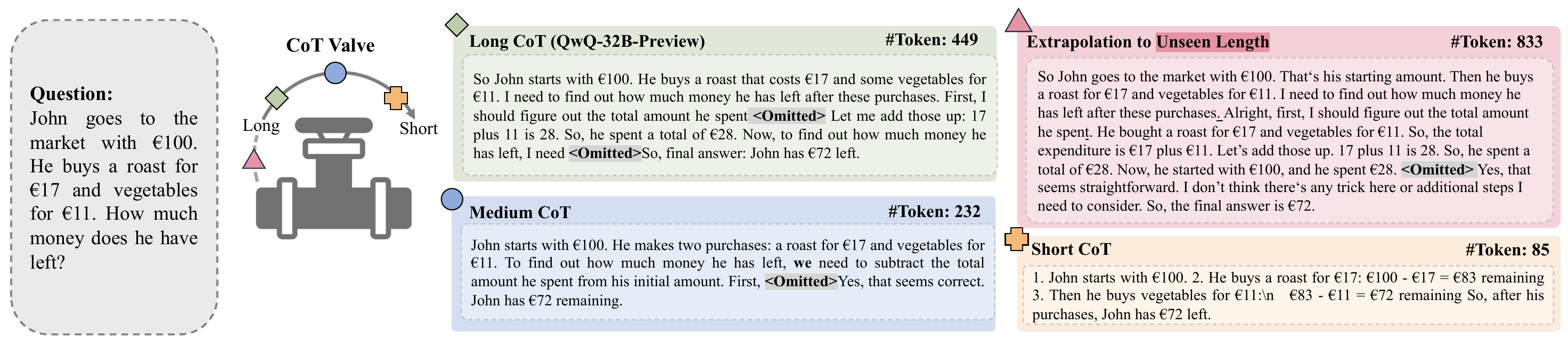}
    \caption{The reasoning model, after the length-compressible CoT tuning, can generate reasoning paths from long to short, leveraging LoRA as a `Valve'. We show one example from our constructed dataset MixChain.}
    \label{fig:teaser}
\end{figure*}

Chain-of-Thought (CoT) reasoning~\citep{wei2022chain} has emerged as a powerful technique for enhancing the reasoning capabilities of large language models~\citep{jaech2024openai,dubey2024llama,abdin2024phi}, particularly in complex tasks such as mathematics and coding ~\citep{sprague2024cot} that require multi-step inference. By simulating the process of human-like thought progression, CoT enables models to break down complex problems into sub-questions, improving accuracy and interpretability~\citep{joshi2023machine}.
Those reasoning abilities have also been tested in different domains, such as image generation~\citep{ma2025inferencetimescalingdiffusionmodels} and visual understanding~\citep{shao2024visual}.

Training reasoning models often involves generating extensive reasoning paths through methods such as sampling~\citep{wang2023selfconsistency}, tree search~\citep{yao2023tree,guan2025rstarmathsmallllmsmaster,zhang2024accessinggpt4levelmathematical} or reinforcement learning~\citep{deepseekai2025deepseekr1incentivizingreasoningcapability} to ultimately reach the correct answer. However, these long chains often incorporate redundant intermediate steps that can be unnecessary or too complex~\citep{lightman2024lets}, and the redundancy in the reasoning paths for training leads to inefficiencies in token usage and increased inference costs. 
However, crafting an optimal reasoning chain that omits extraneous details is challenging due to the limited availability of intermediate rewards to guide the process and human annotations~\citep{zhang2025lessonsdevelopingprocessreward}. Removing some or all of the intermediate steps and then training or distilling the model ~\citep{liu2024can,Yu2024DistillingS2} will degrade the performance. Alternative approaches employ information-theoretic measures~\citep{ton2024understandingchainofthoughtllmsinformation} or identify an "overthinking" solution in QwQ~\citep{qwq-32b-preview} to evaluate the contribution of each sentence to the final answer.


We observe that current reasoning models, such as QwQ~\citep{qwq-32b-preview} and DeepSeek-R1~\citep{deepseekai2025deepseekr1incentivizingreasoningcapability} allocate an excessive number of tokens to simple tasks, while potentially providing insufficient tokens for complex tasks.
\emph{Thus, a long reasoning path is still essential, while maintaining the ability to compress reasoning paths for simpler questions is equally important.}
To solve this, our goal is to fine-tune a model capable of generating both long and short reasoning paths, rather than being restricted to a compressed form.  We offer a new way to control the length of CoT, which we refer to as Length-Compressible Chain-of-Thought Tuning.

A central component of the proposed method is to identify an update direction in the parameter space, which, by manipulating it, acts as increasing or decreasing the length of CoT.
Taking a large step in this direction leads the model to generate a short sequence, while a small step still produces a long and complex reasoning trajectory. 
We choose to incorporate this update direction by LoRA~\cite{hu2022lora}, enabling it to function as an additional branch that facilitates easy modulation of intensity while imposing minimal extra parameters on the model.
We explore methods to identify this direction and demonstrate that it offers superior controllability compared to prompt-based approaches, which enables the generation of short CoT that prompt-based methods are unable to achieve. 
Besides, we observe that the direction can be extrapolated, allowing the reasoning chains to be extended beyond or shortened to lengths unseen in the training set.
Leveraging this compressibility, we construct a dataset that pairs long and short reasoning chains for each question. This dataset is then utilized in two ways: (1) to refine the direction for more precise tuning, and (2) to progressively compress the reasoning path.

We evaluate our method across different types of models, ranging from a pre-trained LLM with little reasoning ability, LLaMA-3.1-8B and LLaMA-3.2-1.5B-Instruct ~\citep{dubey2024llama}, to post-trained reasoning models, QwQ-32B-Preview~\citep{qwq-32b-preview}, and distilled reasoning models, DeepSeek-R1~\citep{deepseekai2025deepseekr1incentivizingreasoningcapability}. 
Our results demonstrate that, with training for one time, our approach enables a model to generate reasoning paths of varying lengths, and we can achieve better results than previous chain compression baselines.
Besides, our study highlights several interesting findings:
(1) Short reasoning paths can sometimes outperform longer ones, underscoring the significance of \methodname\ in enhancing model efficiency.
(2) Not every reasoning chain, despite all leading to the correct final answer, is conducive to model optimization. Excessively long or short chains complicate the distillation of CoT, posing challenges to the model training.


In summary, our contributions are: (1) \textbf{CoT-Valve}: Enables elastic control of length for CoT within the parameter space, allowing a single model to generate CoT from short to long.
(2) \textbf{MixChain Dataset}: A dataset with reasoning paths of varying lengths for each question. 
(3) \textbf{Improved Tuning} \& \textbf{Progressive Compression}: 
Refines the direction-tuning process based on MixChain and introduces progressive compression for inference efficiency.
(4) \textbf{Performance \& Controllability}: Achieves controllable reasoning generation and state-of-the-art results for compressed CoT.

\section{Related Work}
\paragraph{Chain-of-Thought.}
Chain-of-thought~\cite{wei2022chain} reasoning has shown promising progress in recent years, especially the success of OpenAi-O1~\citep{jaech2024openai} and Deepseek-R1 models~\citep{deepseekai2025deepseekr1incentivizingreasoningcapability}. This introduces the test-time scaling law, apart from the traditional scaling law for training~\cite{hoffmann2022trainingcomputeoptimallargelanguage}. Several approaches have been proposed to boost the language model to have better problem-solving abilities, including the model has its self-reasoning abilities~\citep{qwq-32b-preview} or use Best-of-N~\citep{Nakano2021WebGPTBQ}, beam search and Monte Carlo Tree Search~\citep{Kocsis2006BanditBM,guan2025rstar} to search and refine the solution without further finetune the large language models. The outcome reward model and process reward models are also introduced to evaluate the score for the entire solution, especially the final answer~\citep{Cobbe2021TrainingVT} and the 
quality of the reasoning path~\citep{wang-etal-2024-math,luo2025improve}

\paragraph{Chain Compression in reasoning model.} Due to the high computational cost associated with inference in reasoning models, particularly for long-chain reasoning, chain compression has become a critical area of research. \citep{Yu2024DistillingS2} attempts to distill the chain-of-thought into System 1 but fails to observe improvements when intermediate steps are omitted. \citep{deng2024implicit} proposes internalizing reasoning steps within the hidden states of models, while several implicit-based approaches\citep{deng2024explicitcotimplicitcot,hao2024traininglargelanguagemodels,cheng2024compressedchainthoughtefficient} aim to compress token-wise generation by transitioning from language space to hidden space. Other studies focus on skipping intermediate reasoning steps~\citep{liu2024can} or using summarization techniques to generate shorter reasoning chains~\citep{kang2024c3otgeneratingshorterchainofthought}.
Additionally, \cite{chen2024not} addresses the overthinking issue in QwQ~\citep{qwq-32b-preview} and employs SimPO~\citep{meng2024simpo} for optimization. Kimi K1.5~\citep{team2025kimi} proposes merging long-CoT models with short-CoT models in a training-free manner. O1-Pruner~\citep{luo2025o1} adopts reinforcement learning to shorten responses.

\begin{figure*}
    \centering
    \includegraphics[width=\linewidth]{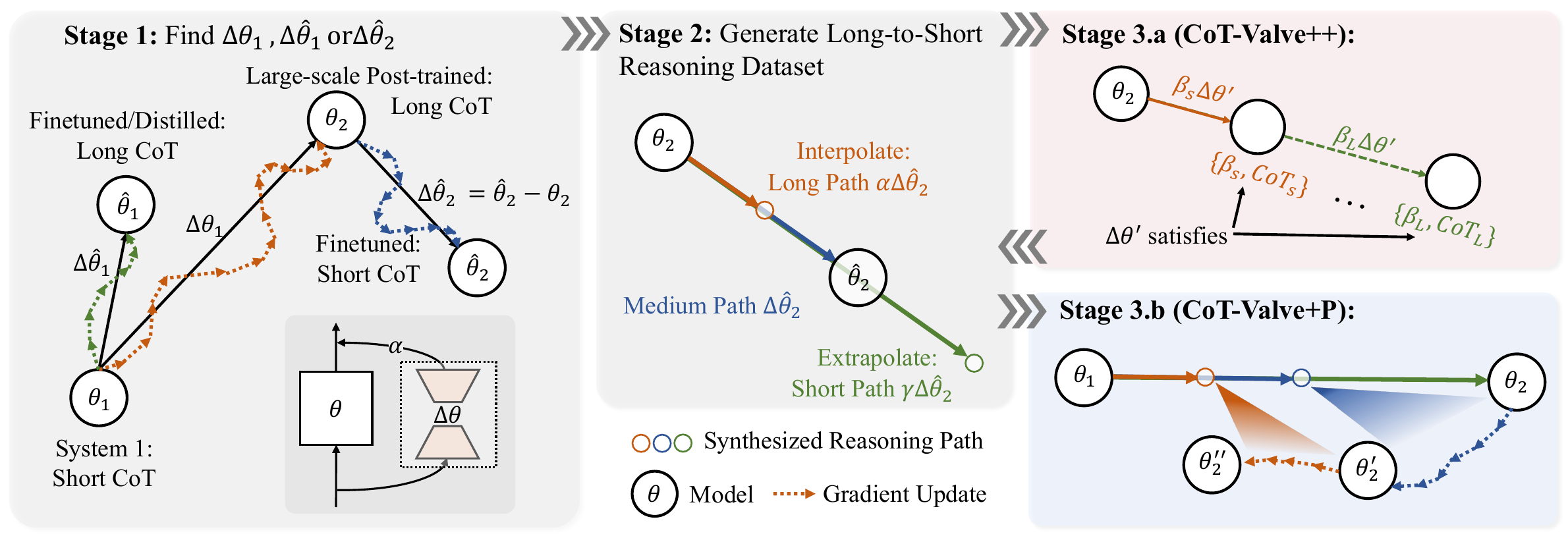}
    \caption{Illustration of \methodname. In Stage 1, we first determine $\Delta\theta$ from distilling or post-training. Then, the trained $\Delta\theta$ is utilized to construct the MixChain dataset. Using this dataset, we can then apply two enhanced training methods to achieve more precise control over reasoning paths, or to shorten the reasoning paths as needed.}
    \label{fig:main}
\end{figure*}

\section{Method}

In this section, we provide an in-depth discussion of our method. Section \ref{sec:cut} introduces a simple yet effective approach that enables a single tuning process to generate models with CoT with different lengths. This stage also serves as an initial step for subsequent refinements. Next, in Section \ref{sec:construct_dataset}, we explore multiple scenarios in which we can apply \methodname\ to construct the dataset MixChain. In Section \ref{sec:improve_cut}, we propose several advanced methods that take advantage of long-to-short datasets to improve precision and control over the generated reasoning paths in compressible fine-tuning.

\subsection{Length-Compressible CoT Tuning} \label{sec:cut}

Our primary objective is to achieve a new way to control the length of reasoning paths after training a reasoning model. Existing approaches, such as prompt-based control, explicitly define sequence length in the prompt \cite{han2024token} or utilize summary tokens \cite{ding2024break} for guidance. However, these methods offer only limited control over the length of CoT generated. For instance, requesting a sequence of less than 20 tokens may result in the model generating over 350 tokens (see Table \ref{tbl:prompt-control} in the Appendix), and these methods struggle to produce answers with very short lengths. To address these limitations, we introduce \methodname\ for training one model but can adjust the length of reasoning paths.

Consider a reasoning model defined by the parameter $\theta$. For a given question 
$q$ in the dataset $\mathcal{D}$, the probability of generating an answer $a$ and its reasoning thoughts $\{t_i\}_{i=1}^n$ given the question $q$ can be described by:
\begin{equation}
    p\left(a \mid t_1, \ldots, t_n, q; \theta \right) \prod_{i=1}^n p\left(t_i \mid t_{<i}, q; \theta \right) 
\end{equation}
where $\{t_i\}_{i=1}^n$ might include errors or unnecessary details. With short synthesized or human-annotated explanations $\{t_i\}_{i=1}^m$ with $m<n$, the training objective is to adjust the parameter in such a way that the chain is shortened while still yielding the correct answer:
\begin{align}
\max_{\Delta \theta} \mathbb{E}_{(q, a) \sim \mathcal{D}}  & p\left(a \mid t_1, \ldots, t_m, q; \theta + \Delta\theta\right) \nonumber \\
& \prod_{i=1}^m p\left(t_i \mid t_{<i}, q; \theta + \Delta\theta \right) 
\end{align}
and $\Delta\theta$ denotes the change in the parameter space that steers the model towards generating a more concise chain. 

Since the model, with and without $\Delta\theta$, outputs the same final answer, $\Delta \theta$ can be interpreted as a task vector~\citep{ilharco2022editing}. The task here is to control the length of the CoT, provided that the only difference in the training set lies in intermediate reasoning steps $\{t_i\}_{i=1}^n$. Those reasoning paths are different in length but ultimately lead to the same final answer. Thus, we can control the task vector to achieve the goal of adjusting the length of CoT.
$\Delta \theta$ is designed within a parameter-efficient space, functioning as an external branch for inference that incurs minimal overhead. Controlling this external branch enables the manipulation of the length of the reasoning path.

\paragraph{Task Arithmetic: Interpolation and Extrapolation of $\Delta\theta$.}
To manipulate this update within the parameter space, we can control the magnitude of a $\Delta\theta$ as an arithmetic operation. We use two primary operations on $\Delta\theta$ here: interpolation and extrapolation. 
Let $\alpha$ denote the magnitude of $\Delta\theta$ for LoRA. When $\alpha$ falls within the range of (0,1), the model smoothly transitions between longer and shorter reasoning paths, similar to weight interpolation between two models~\citep{frankle2020linear,team2025kimi}.
When $\alpha>1$, extrapolation is introduced, further shortening the reasoning path beyond what was observed during training. This enables an exploration of the minimal reasoning length required to arrive at a given answer. Thus, by adjusting $\alpha$ at inference, we can modulate the model's behavior, with each value of $\alpha$ corresponding to different CoT lengths.


\paragraph{Application}
Unlike prompt-based approaches that can only regulate the overall length of the reasoning process using prompt words, $\Delta\theta$ provides finer granularity control.
$\Delta\theta$ is served in the external parameter space. This allows for greater flexibility in adjusting the reasoning trajectory. Specifically, it facilitates the selective retention of long-chain reasoning in certain thoughts while applying stronger compression to simpler reasoning segments. As a result, reductions in chain length can be localized to specific portions of the inference process rather than being uniformly applied across the entire reasoning path. We remain the design of this segment selection in future work.

\subsection{Construct the MixChain Dataset} \label{sec:construct_dataset}
A crucial thing for the above process is the construction of the training dataset, especially the reasoning chain $\{t_i\}_{i=1}^n$. 
To have reasoning chains with different lengths, previous approaches rely on multiple rounds of sampling, selecting reasoning paths under different random seeds, or using some handcrafted way to remove parts of the answer~\citep{chen2024not}. 

We introduce MixChain, a dataset inherently generated by our method that contains reasoning paths of varying lengths. This dataset is structured such that each question is associated with multiple reasoning paths, with lengths progressively decreasing from long to short. By simply adjusting the parameter $\alpha$, our approach avoids the need for repeated sampling and achieves this diverse set of reasoning paths. In contrast to multi-sampling techniques, MixChain enables a more reliable and consistent generation of shorter reasoning paths while simultaneously capturing a spectrum of reasoning lengths.
To construct MixChain, we consider two possible scenarios:
\begin{itemize}
    \item If a well-annotated dataset with human-labeled solutions is available, such as GSM8K~\citep{cobbe2021trainingverifierssolvemath} or PRM800k~\citep{lightman2024lets}, it can be leveraged to fine-tune the model for generating shorter reasoning chains as a cold start ($\theta_1 \rightarrow \tilde{\theta}_1$ and $\theta_2 \rightarrow \tilde{\theta}_2$ in Figure \ref{fig:main}). 
    \item In the absence of a dataset containing explicit reasoning paths, or when only final answers are available without full explanations, training solely on final answers is unlikely to enable the model to generate reasoning steps. To address this limitation, we propose an alternative method for constructing MixChain. Specifically, we leverage an existing base LLM (e.g., LLaMA-3.1-8B or Qwen-32B-Instruct) as $\theta_1$ and use its corresponding reasoning model (e.g., DeepSeek-R1-Distill-Llama-8B or QwQ-Preview) to derive $\Delta \theta$. The parameter update between these models serves as a form of linear interpolation, enabling the transition from $\theta_1$ to $\theta_2$. This transition is then used to construct the dataset, as illustrated in Figure \ref{fig:main}, where the parameter shift is represented by $\theta_1 \rightarrow \theta_2$.
\end{itemize}

\subsection{Improved Tuning for CoT-Valve} \label{sec:improve_cut}
In this section, we present two enhanced variants of \methodname: one aimed at achieving improved controllability and the other focused on optimizing the compression ratio of the reasoning paths.

\paragraph{A More Precise CoT-Valve Paradigm: \methodname++.} In the previously proposed CoT-Valve framework, the training process only constrained $\Delta\theta$ to satisfy the final objective with $\alpha=1$. However, during inference, we expect all positions along this direction to exhibit reasoning trajectories of varying lengths. This leads to the inconsistency between training and inference. With MixChain, we can explicitly incorporate this requirement during training by introducing an additional constraint, ensuring that the model can adapt to reasoning chains of different lengths across all positions in this direction. For each training sample, in addition to the question, answer, and solution, we have introduced a normalized term $\beta$, which represents the factor for the length of the reasoning path. 
Under this dataset, our training objective is modified to find a parameter update $\Delta\theta^{\prime}$ such that it satisfies:
\begin{align}
    \max_{\Delta\theta^{'}} \mathbb{E}_{(q, a) \sim \mathcal{D'}}& p\left(a \mid t_{<m}, q; \theta + \beta \Delta\theta^{'}\right) \nonumber \\
    &\prod_{i=1}^{m} p(t_i | t_{<i}, q; \theta + \beta \Delta\theta^{'})
\end{align}
Where $\mathcal{D^\prime}$ is the Mixchain dataset. Each sample consists of the question $q$, the answer $a$, the solution $\{t_i\}_{i=1}^m$ and $\beta$, where $\beta$ is calculated as:
\begin{equation}
    \beta = 1 - \frac{m - m_{min}}{m_{max} - m_{min}}
\end{equation}
Here, $m_{min}$ and $m_{max}$ is the length of the shortest solution and longest solution for this question. Based on synthetic samples, we introduce additional constraints that enable us to better identify the updated parameter $\Delta\theta^{'}$, facilitating more precise compressibility and controllability.

\paragraph{Progressive Chain Compression: \methodname+P.}
The structure of MixChain, which features progressively shorter reasoning paths for each question, facilitates a progressive chain-length compression strategy. This approach is similar to iterative pruning in model compression~\cite{molchanov2016pruning}. In this process, the model is trained with a shorter reasoning path from the dataset at each iteration, rather than training directly with the shortest reasoning CoT. This gradual compression method allows the model to progressively reduce the length of its reasoning paths.

\section{Experiments}

\begin{figure*}[t!]
  \centering
  \begin{subfigure}{0.32\textwidth}
    \centering
    \includegraphics[width=\textwidth]{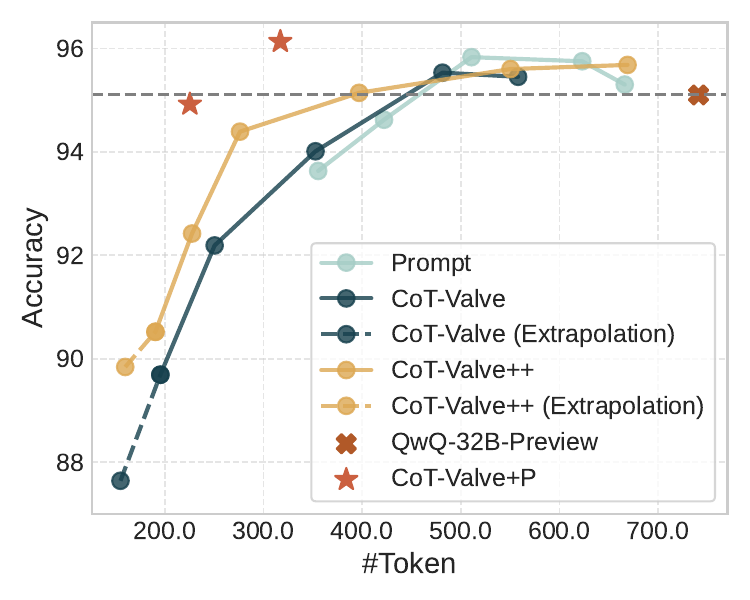}
    \caption{GSM8K, QwQ-32B-Preview}
    \label{fig:qwq-gsm8k}
  \end{subfigure}
  \hfill
  \begin{subfigure}{0.32\textwidth}
    \centering
    \includegraphics[width=\textwidth]{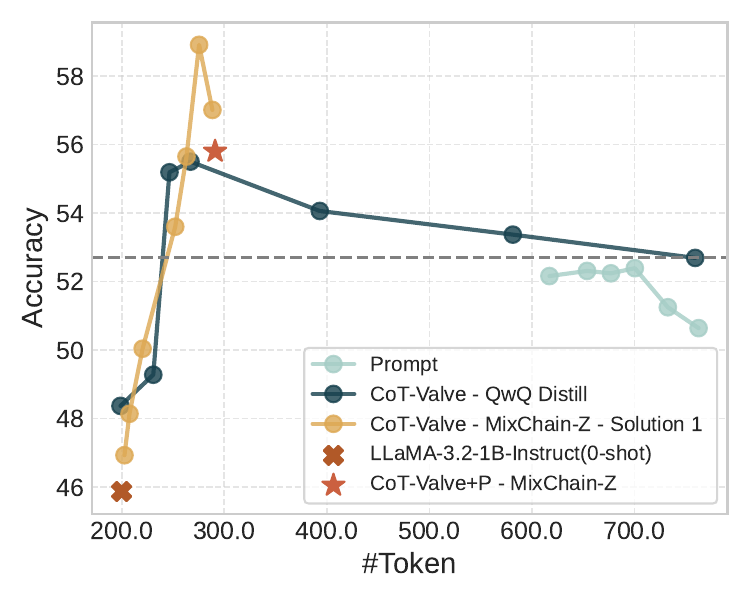}
    \caption{GSM8K, Llama-3.2-1B-Instruct}
    \label{fig:llama1b-gsm8k}
  \end{subfigure}
  \hfill
  \begin{subfigure}{0.33\textwidth}
    \centering
    \includegraphics[width=\textwidth]{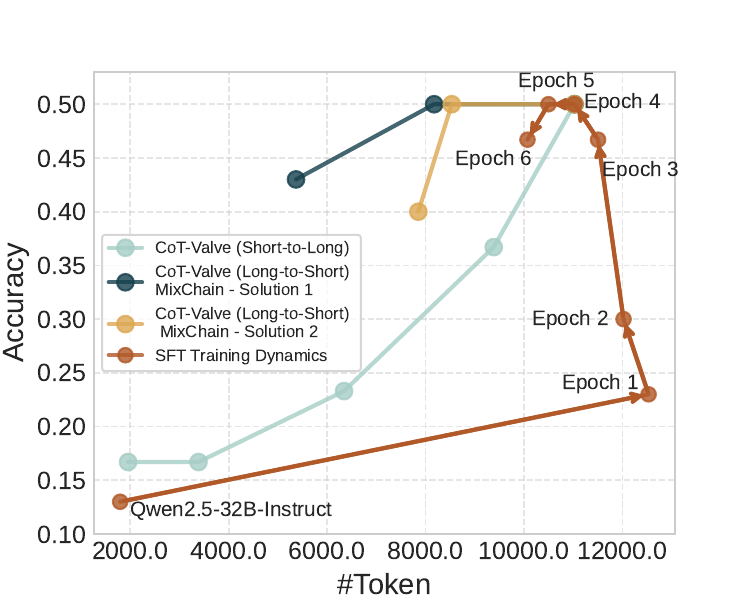}
    \caption{AIME, Qwen2.5-32B-I w/ LIMO}
    \label{fig:limo-aime}
  \end{subfigure}
  \caption{Token length and accuracy for different methods, datasets and reasoning models. Points connected by curves in (a) and (b) represent results from one model.}
  \label{fig:exp_main}
\end{figure*}

\subsection{Experimental Setup}
\paragraph{Models.}
We evaluate our method under several models: QwQ-32B-Preview~\citep{qwq-32b-preview}, DeepSeek-R1-Distill-Llama-8B~\citep{deepseekai2025deepseekr1incentivizingreasoningcapability}, LLaMA-3.1-8B~\citep{dubey2024llama}, LLaMA-3.2-1B~\citep{dubey2024llama} and Qwen-32B-Instruct~\cite{qwen2.5} with LIMO~\citep{ye2025limo}. We tested different scenarios for \methodname: 
\begin{itemize}
    \item \textbf{(Long to Short CoT)} For QwQ-32B-Preview (QwQ for abbreviation) and DeepSeek-R1-Distill-Llama-8B (R1-Distill), we used our method to control and compress the length of the reasoning chain.
    \item \textbf{(Short to Long CoT)} For LLaMA-3.1-8B and LLaMA-3.2-1B-Instruct, we applied our method to distill reasoning abilities from QwQ-32B-Preview and incorporated \methodname\ in the distillation process. 
    \item \textbf{(Short-Long-Short CoT)} We tested another setting to first post-train a short-CoT LLM, Qwen-2.5-32B-Instruct~\citep{qwen2.5}, to generate Long CoT and then compress it to Short CoT. \methodname\ can be applied in both two stages.
\end{itemize}

\paragraph{Metrics.}
We report both accuracy and the number of tokens in the answer for each experiment. Given the trade-off between reasoning path length, model size, and performance, we use a new metric, Accuracy per Computation Unit(ACU), to better capture this balance and evaluate model efficiency. It is defined as:
\begin{equation}
    \text{ACU} = \frac{\text{Accuracy}}{\text{\#Params} \times \text{\#Tokens}}
\end{equation}
Since the ACU value typically falls within the range of $10^{-5}$ to $10^{-2}$, we report it in units of $10^{2}$ for improved readability.

\paragraph{Training and Evaluation.} For training the model, we use LoRA~\citep{hu2022lora} in most of our experiments, except in the experiment for LIMO on Qwen-2.5-32B-Instruct we use full parameter fine-tuning. We also show the results using DoRA~\citep{DoRA} in the Appendix.
The hyper-parameters for each experiment are shown in Appendix \ref{apx:implementation_details}.
We select two math datasets to evaluate the performance, for one easy math dataset, GSM8K~\citep{cobbe2021trainingverifierssolvemath} and one hard math dataset, AIME24. 

\subsection{Datasets} \label{sec:datasets}
We find in our experiments that the quality of the solution is important to the performance, even if all the human-annotated solutions or synthesized solutions reach the final answer. 
In our experiments, we use the question from the train set of GSM8K, the math split of PRM800K or the question from LIMO, and we employ three types of datasets with those questions in our experiments:

\begin{itemize}
    \item Ground-truth Dataset: The dataset provides a human-annotated or model-synthesized solution. We use this as the cold start.
    \item MixChain from cold-start (MixChain-C): After taking the ground-truth dataset to train the model, we can get the first model to generate solutions from short to long. Then we use it to generate the dataset.
    \item MixChain from zero-shot (MixChain-Z):  We employ \methodname\ between a reasoning model ($\theta_2$) and a base LLM ($\theta_1$) to generate the solutions. 
\end{itemize}
For each dataset, we filter out all the solutions with incorrect answers. We show the statistics of the dataset in Table \ref{tbl:dataset} in the Appendix.

\begin{table}[t!]
    \begin{center}
    \resizebox{\linewidth}{!}{
    \begin{tabular}{l|c|rr}
        \toprule
        Method  & Accuracy & \#Token & ACU{ $\uparrow$} \\
        \midrule
        Llama-3.3-70B-Instruct &  92.6 & 235.4 & 0.56 \\
        Llama-3.1-405B-Instruct &  95.6 & 186.7 & 0.13  \\
        Qwen2.5-32B-Instruct & 93.1 & 269.3 & 1.09  \\
        Qwen2.5-Math-72B-Instruct & 95.8 & 312.1 & 0.43 \\
        QwQ-32B-Preview  & 95.1 & 741.1 & 0.40\\
        \midrule
        Prompt \cite{han2024token} & 93.6 & 355.5 & 0.82 \\
        Prompt \cite{ding2024break} & 95.5 & 617.7 & 0.48 \\
        \midrule 
        \multicolumn{4}{c}{In-domain Train Set: GSM8K} \\
        \midrule 
        \methodname\ - Ground-Truth & 94.0 & 352.8 & 0.83 \\
        \methodname++ - MixChain-C & 94.4 & 276.3 & 1.07 \\
        \methodname+P - MixChain-Z & 96.1 & 317.1 & 0.95 \\
        \methodname+P - MixChain-Z & 94.9 & 225.5 & 1.32 \\
        \midrule
        \multicolumn{4}{c}{Out-of-Domain Train Set: PRM12K } \\
        \midrule
        Overthink\cite{chen2024not} - SFT & 94.8 & 749.5 & 0.40 \\
        Overthink\cite{chen2024not} - SimPO & 94.8 & 326.2 & 0.91 \\
        O1-Pruner\cite{luo2025o1} - SFT & 95.7 & 717 & 0.42\\
        O1-Pruner\cite{luo2025o1} & 96.5 & 534 & 0.56 \\
        \methodname+P - MixChain-Z & 95.4 & 288.5 & 1.03\\
        \bottomrule
    \end{tabular}
    }
    \caption{Results of QwQ-32B-Preview on GSM8K. Values of ACU are scaled by $10^{2}$ for readability. We list the dataset we use after the method name.}
    \label{tbl:qwq-gsm8k}
    \end{center}
\end{table}

\begin{table}[t]
    \begin{center}
    \resizebox{\linewidth}{!}{
    \begin{tabular}{l|c|rr}
        \toprule
        Method  & AIME24 & \#Token & ACU$\uparrow$\\
        \midrule
        Qwen2.5-32B-Instruct  &  4/30 & 1794.2 & 0.023 \\
        Qwen2.5-Math-72B-Instruct &  7/30 & 1204.5 & 0.061 \\
        Gemini-Flash-Thinking~\citep{team2023gemini} & 15/30 & 10810.5 & - \\
        \midrule
        \multicolumn{4}{c}{QwQ-32B-Preview.Train set: GSM8K} \\
        \midrule
        QwQ-32B-Preview & 14/30 & 6827.3 & 0.021 \\
        Prompt \cite{han2024token} & 13/30 & 6102.5 & 0.022  \\
        Prompt \cite{ding2024break} & 13/30 & 5562.3 & 0.024\\
        Overthink \cite{chen2024not}  & 13/30 & 5154.5 & 0.026 \\
        \midrule
        \methodname\ - GSM8K & 14/30 & 5975.0 & 0.024 \\
        \methodname++ - MixChain-C & 13/30 & 5360.5 & 0.025  \\
        \methodname+P - MixChain-Z & 13/30 & 4629.6 & 0.029 \\
        \midrule
        \multicolumn{4}{c}{Qwen-32B-Instruct. Train set: LIMO}\\
        \midrule
        Qwen-32B-LIMO & 15/30 & 10498.2 & 0.015 \\
        CoT-Valve & 11/30 & 6365.2 & 0.018 \\
        SFT - MixChain - Solution 1 & 13/30 &  5368.0 & 0.025  \\
        CoT-Valve - MixChain - Solution 1 & 15/30 &  8174.8 & 0.019  \\
        \bottomrule
    \end{tabular}
    }
    \caption{Results of QwQ-32B-Preview and Qwen-32B-Instruct w/ LIMO on AIME 24.}
    \label{tbl:qwq-aime}
    \vspace{-3mm}
    \end{center}   
\end{table}

\begin{table}[t]
    \begin{center}
    \resizebox{\linewidth}{!}{
    \begin{tabular}{l|rr|rr}
        \toprule
        & \multicolumn{2}{c|}{GSM8k} & \multicolumn{2}{c}{AIME24}\\
        Model  & Acc & \#Token & Acc & \# Token \\
        \midrule
        Llama-3.1-8B (0-shot) & 15.7 & 915.0 & 0/30 & 1517.6 \\
        R1-Distill-Llama-8B & 87.1 & 1636.6 & 14/30 & 12359.9\\
        \midrule 
        \methodname\ & 87.3 & 1315.2 & 6/30 & 7410.5\\
        \methodname+P - MixChain-Z & 84.0 & 755.2 & 11/30 & 9039.0 \\
        \bottomrule
    \end{tabular}
    }
    \caption{Result of DeepSeek-R1-Distill-Llama-8B. }
    \label{tbl:deepseek-r1}
    \end{center}
\end{table}

\subsection{From Long-CoT to Short-CoT.}
\paragraph{Controllable Results.} We illustrate the result in Figure \ref{fig:qwq-gsm8k}.
First, using ground-truth samples as a cold start, we develop a model capable of generating reasoning paths of various lengths, as demonstrated in `CoT-Valve' in Figure \ref{fig:qwq-gsm8k}. CoT-Valve already matches the performance of prompt-based control but can generate shorter reasoning chains. We then extrapolate $\Delta\theta$ to produce even shorter reasoning paths. Then, building on MixChain-C from this first model, we conduct further training by CoT-Valve++. CoT-Valve++ substantially surpasses the baseline and shows greater generalization capabilities in cases of extrapolation.


\paragraph{Compression Results.} We evaluated our method against previous chain compression approaches, with the results detailed in Table \ref{tbl:qwq-gsm8k}, Table \ref{tbl:qwq-aime}, and Table \ref{tbl:deepseek-r1}. For GSM8K, we adhered to the baseline setup to train with PRM12K. Utilizing progressive compression, our method surpassed the baseline by producing shorter reasoning paths and improved performance.

We also report experimental results on AIME, where the model was trained using MixChain-Z derived from GSM8K. To minimize the impact of randomness on performance, we employed greedy decoding in our AIME experiments. Compared to the baseline ~\citep{chen2024not}, our method reduced the token count from 5155 to 4630 while maintaining the same accuracy, despite being trained on an easier dataset.

\begin{table}[t]
    \begin{center}
    \resizebox{\linewidth}{!}{
    \begin{tabular}{l|c|rr}
        \toprule
        Method & Accuracy & \#Tokens & ACU$\uparrow$  \\
        \midrule
        LLaMA-3.2-1B-Instruct(8-shot) &  45.9 & 104.3 & 44.008\\
        LLaMA-3.2-1B-Instruct(0-shot) &  45.9 & 199.8 & 22.973\\
        \midrule
        SFT-Full Finetune - GSM8k & 46.1 & 139.4 & 33.070\\
        SFT - GSM8k & 43.8 & 137.7 & 31.808\\
        Prompt & 46.7 & 209.9 & 22.249\\
        \midrule 
        SFT - QwQ Distill & 52.7 & 759.3 & 6.941\\
        \methodname\ - QwQ Distill & 55.5 & 267.0 & 20.786\\
        \methodname+P - MixChain-Z & 55.8 & 291.0 & 19.175\\
        SFT - MixChain-Z - Solution 1 & 57.0 & 288.4 & 19.764\\
        \methodname\ - MixChain-Z - Solution 1 & 58.9 & 275.4 & 21.387\\
        \bottomrule
    \end{tabular}
    }
    \caption{Results on LLaMA-3-2-1B-Instruct. We report the result of Flexible Match here. QwQ Distill means we use QwQ to synthesize the solution and distill it.}
    \label{tbl:llama1b-gsm8k}
    \end{center}
\end{table}

\begin{table}[t]
    \begin{center}
    \resizebox{\linewidth}{!}{
    \begin{tabular}{l|c|rr}
        \toprule
        Method & Accuracy & \#Tokens & ACU$\uparrow$ \\
        \midrule
        LLaMA-3.1-8B (8-shot)  & 56.9 & 282.1 & 2.521\\
        LLaMA-3.1-8B (0-shot) & 15.7 & 915.0 & 0.214\\
        \midrule 
        SFT-LoRA - GSM8k & 59.0  & 191.9 & 3.843\\
        \midrule
        SFT-LoRA - QwQ Distill & 76.3 & 644.8 & 1.479\\
        \methodname\ - QwQ Distill & 77.5 & 569.8 & 1.700\\ 
        \methodname+P - MixChain-Z & 77.1 & 371.2 & 2.596\\
        CoT-Valve + MixChain-Z - Solution 1& 75.7 & 264.1 & 3.583\\
        
        \bottomrule
    \end{tabular}
    }
    \caption{Result on LLaMA-3.1-8B. We report the result of Strict Match here.}
    \vspace{-3mm}
    \label{tbl:llama8b-gsm8k}
    \end{center}
\end{table}

\subsection{From Short-CoT to Long-CoT \& Short-Long-Short CoT} 
Our method can also be applied if a short-CoT model is distilled or post-trained to be a Long-CoT model.
The results are shown in Figure \ref{fig:llama1b-gsm8k}, Table \ref{tbl:llama1b-gsm8k} and Table \ref{tbl:llama8b-gsm8k}. 
We found that CoT-Valve can also effectively control the length of the chains in this setting. Notably, we observed that shorter chains could achieve higher accuracy on GSM8K. Moreover, if the model is trained using the MixChain-Z dataset, the results are significantly better, whether using CoT-Valve (55.5 to 58.9) or just simply SFT (52.7 to 57.0). 
Additionally, after training a long-chain model, we can employ the MixChain dataset to reduce the length of its reasoning chains further. As illustrated in Figure \ref{fig:limo-aime}, the results suggest that initially training the chains to be long and subsequently compressing them to be shorter (Results with Long-to-Short) can yield better performance than directly using CoT-Valve in the short-to-long stage (Results with Short-to-Long). This demonstrates significant potential for compressing the reasoning chains. We can also surpass the result of Gemini-Flash-Thinking, with the same accuracy but fewer tokens (10810.5 v.s. 8174.8)

\paragraph{Training dynamics does not have the same effect as \methodname.} 
We also explore whether intermediate training steps can achieve similar effects. As depicted in Figure \ref{fig:limo-aime}, during the early training phases, the length of the CoT increases but does not correspond with the same rapid improvement in performance. As training progresses, the token length begins to decrease while performance improves. \methodname\ exhibits a distinct pattern, smoothly bridging the gap between the length of CoT and performance.

\subsection{Observations}
Based on the results from LLaMA-3.1-8B, LLaMA-3.2-1.5B, QwQ, DeepSeek-R1-Distill-Llama-8B and Qwen2.5-32B-Instruct with LIMO, we summarize the following observations:
\begin{itemize}
    \item \textbf{Longer reasoning chains are not always the best on simple datasets}. 
    Across nearly all models, we find that those directly trained on long CoT data typically do not show the best performance. These models often underperform compared to those generated through CoT-Valve, which results in shorter but more accurate reasoning chains. This trend is particularly pronounced in smaller models. For instance, in the LLaMA-3.2-1B model, training on QwQ synthesized data yields an accuracy of 52.69 with 759.3 tokens. However, using \methodname, we can achieve an accuracy of 55.50 with only 267.0 tokens. However, we do not observe this phenomenon in more complex datasets, indicating that while the reasoning model may be redundant for simple datasets, it still requires test-time scaling to effectively handle complex datasets.
    \item \textbf{Some reasoning chains are difficult for the model to learn, especially for small LLMs.} 
    We fine-tuned LLaMA-3.2-1B-Instruct using only one solution from MixChain, where all solutions lead to the same final answer but involve different intermediate reasoning steps. The results, presented in Table \ref{tbl:solution_distill}, indicate that neither the shortest nor the longest chains are optimal for learning. Instead, the model most effectively learns from moderately short chains, achieving the highest accuracy while maintaining a relatively low token count. This phenomenon is particularly evident in smaller models, but it is not observed in larger models. We believe this could be beneficial for the distillation of CoT in small LLMs.
\end{itemize}

\begin{table}[t]
    \centering
    \resizebox{\linewidth}{!}{
    \begin{tabular}{c|c|cc}
        \toprule
        Solution & Solution Length & Accuracy & \#Token \\
        \midrule
        Ground-Truth (Solution 0) & 116.0 & 43.8  & 139.4 \\
        Solution 1	& 279.6 & 57.0	& 288.4 \\
        Solution 2	& 310.7 & 55.1	& 330.0 \\
        Solution 3	& 386.7 & 56.5	& 414.6 \\
        Solution 4	& 497.2 & 52.5	& 558.3 \\
        \bottomrule
    \end{tabular}
    }
    \caption{Train LLaMA-3.2-1B-Instruct with solutions in MixChain-Z of different lengths on GSM8K.}
    \label{tbl:solution_distill}
\end{table}

\subsection{Analysis}
\paragraph{Ablation on Progressive Compression.}
Table \ref{tbl:progressive} demonstrates the effect of progressive compression.
We compare two settings: training directly with the ground-truth solution for five epochs 
and applying progressive compression for five epochs in total, with the final epoch using the ground-truth data.
Our results show that progressive compression significantly improves the performance of short CoT (from 92.19 to 94.92). 
For each turn, progressive compression gradually reduces the token number while maintaining accuracy.

\paragraph{CoT-Valve achieves shorter chains compared to prompt control} 
We also present in Table \ref{tbl:short-prompt-control} the shortest chain achieved by our method and compare these with those obtained using prompt control. Our method outperforms prompt control methods at shorter chain lengths. Additionally, we explored the limits of chain length for both methods and found that our approach can generate substantially shorter chains than what can be achieved through prompt control.

\begin{table}[t]
    \begin{center}
    \resizebox{\linewidth}{!}{
    \begin{tabular}{l|cc|cc|c}
        \toprule
        Solution Used & \#Epoch & \#Samples & Accuracy & \#Tokens & ACU$\uparrow$   \\
        \midrule 
        - & - & - & 95.07 & 741.1 & 0.40 \\
        \midrule
        4 & 1 &  6.8k & 95.68 & 597.3 & 0.50 \\
        4+3 & 1 &  13.7k & 94.84 & 458.4 & 0.65 \\
        4+3+2 & 1 &  20.5k & 94.84 & 339.9 & 0.87 \\
        4+3+2+1 & 1 & 27.4k & 96.13 & 317.1 & 0.95  \\
        4+3+2+1+0 & 1 & 34.2k & 94.92 & 225.5 & 1.32 \\
        \midrule 
        0 & 5 & 37.4k & 92.19 & 250.5 & 1.15 \\
        \bottomrule
    \end{tabular}
    }
    \caption{Ablation of Progressive Compression on QwQ. Here, solution 0 is the human-annotated solution from the original dataset. }
    \label{tbl:progressive}
    \end{center}
\end{table}

\begin{table}[t!]
    \begin{center}
    \resizebox{\linewidth}{!}{
    \begin{tabular}{c|cc|cc}
        \toprule
        &\multicolumn{2}{c|}{QwQ-32B-Preview} & \multicolumn{2}{c}{Llama-3.2-1B-I} \\
        Method & Acc & \#Token & Acc & \#Token\\
        \midrule 
        Prompt (Shortest)& 93.6 & 355.5 & 52.5 & 621.0 \\
        Ours (Best)  & 94.4 & 276.3 & 55.5 & 267.0\\
        Ours (Shortest) & 87.5 & 133.8 & 50.4 & 247.0 \\
        \bottomrule
    \end{tabular}
    }
    \caption{\methodname\ can achieve shorter chains than prompts with better performance.}
    \label{tbl:short-prompt-control}
    \vspace{-5mm}
    \end{center}
\end{table}

\section{Conclusion}
In this paper, we propose a method that enables a model to generate reasoning chains of varying lengths instead of the prompt control. Based on this approach, we construct a dataset containing both long and short reasoning chains to further enhance controllability and compression efficiency. Experimental results demonstrate the effectiveness of our method in dynamic reasoning chain control and the compression of  CoT. Future research can further explore finer-grained control strategies to improve reasoning efficiency and model controllability.



\bibliography{custom}

\clearpage

\appendix

\section{Implementation Details}
\label{apx:implementation_details}

\subsection{Evaluation Metric.} For experiments on LLaMA, we use lm-eval-harness\footnote{https://github.com/EleutherAI/lm-evaluation-harness} to evaluate the model performance. 
For LLaMA-3.1-8B, we report the strict matching metric due to observed repetition in the model’s responses, which causes the flexible match to extract incorrect numerical values.
For LLaMA-3.2-1B-Instruct, we report results using the flexible match metric.
For QwQ-32B-Preview, DeepSeek-R1-Distill-Llama-8B and Qwen-2.5B-LIMO, we first extract the result enclosed within \textbackslash boxed\{\}. If no such boxed answer is found, we default to using the last digit in the response as the final answer. 

\subsection{Training Setting.}
\paragraph{LLaMA-3.1-8B} The model is trained using eight A5000 24GB GPUs. We set the batch size to 64 and the peak learning rate to 4e-5, following a cosine decay schedule. A weight decay of 0.01 is applied. For the progressive chain compression experiment, we train the model for two epochs with each type of solution. For all other experiments, we train for a maximum of eight epochs. For LoRA, the rank is set to 32, and the lora\_alpha for training is set to 64. During inference, the maximum number of tokens is set to 2048.

\paragraph{LLaMA-3.2-1B-Instruct} The model is trained using 8 A5000 24GB GPUs. We set the batch size to 8 for the CoT-Valve experiment and 64 for all other experiments. The peak learning rate is 4e-5, following a cosine decay schedule, except for the SFT - GSM8K experiment, where the peak learning rate is 1e-5. A weight decay of 0.01 is applied. For the CoT-Valve and SFT-Full Finetune - GSM8k experiment, we train for a maximum of four and six epochs, respectively. For the progressive chain compression experiment, we train the model for two epochs with each type of solution. For all other experiments, training is conducted for up to 8 epochs. For LoRA, the rank is set to 32, and the lora\_alpha for training is set to 64. During inference, the maximum number of tokens is set to 2048.

\paragraph{QwQ-32B-Preview.} 
The model is trained on two H100-80G GPUs. We set the batch size to 64 and trained for a maximum of five epochs. The learning rate is 1e-5, with a weight decay of 0.01 applied during training. For LoRA, the rank is set to 2, and the lora\_alpha for training is set to 8. During inference, we set the maximum token to be 4192 for GSM8K and the maximum token as 8192 for AIME correspondingly.

\paragraph{DeepSeek-R1-Distill-Llama-8B.}
Our experiment on DeepSeek-R1-Distill-Llama-8B\footnote{https://huggingface.co/deepseek-ai/DeepSeek-R1-Distill-Llama-8B} is conducted using the MixChain-zero-shot-GSM8K dataset. The batch size is set to 128, and training is performed for a maximum of five epochs. To ensure that the inference process successfully generates the final answer, we set the maximum token limit to 30K.

\paragraph{Qwen2.5-32B-LIMO.} We fine-tuned Qwen-32B-Instruct using LIMO, training on four H100 GPUs for 10 epochs with a batch size of 4 and a maximum sequence length of 16K. The learning rate was set to 5e-6.
We define Qwen-32B-Instruct as $\theta_0$ and the trained model as $\theta_1$, treating the update direction between them as $\Delta \theta$. By adjusting $\alpha$, we generated the MixChain-C-LIMO dataset, which includes two solutions: solution 1 ($\alpha$=0.8) and solution 0 ($\alpha$=0.6).

Based on this, we further trained $\theta_2$ for 5 epochs with a batch size of 32, a learning rate of 5e-6, and a weight decay of 0.01, obtaining the results of MixChain-Solution 0 in Table \ref{tbl:qwq-aime}. This model can be further refined through CoT-Valve (Results: CoT-Valve + MixChain - Solution 0). Unlike previous experiments, we applied full fine-tuning instead of LoRA. The maximum generated sequence length in this experiment was 15K.
  
\subsection{Dataset Explanation}
As detailed in Section \ref{sec:datasets}, we constructed two types of datasets: MixChain-C and MixChain-Z. The statistics for the datasets are shown in \ref{tbl:dataset}. For these datasets, we select $\alpha$ values ranging from [0.6, 0.8] for LIMO and [0.2, 0.4, 0.6, 0.8] for other datasets, ensuring all incorrect responses are excluded.

For MixChain-Z, while the training transition from $\theta_1$ to $\theta_2$ remains a black box, we can still identify numerous model pairs such as Qwen-32B-Instruct $\rightarrow$ QwQ-32B-Preview, and LLaMA-3.1-8B $\rightarrow$ R1-Distill-Llama-8B, as documented in the technical report. We find that the performance of the base model significantly influences the quality of the dataset.

\begin{table}[t]
    \begin{center}
    \resizebox{\linewidth}{!}{
    \begin{tabular}{c|c|cc}
        \toprule
        Dataset & Solution Index & \#Samples  & \#Avg Token  \\
        \midrule
        \multicolumn{4}{c}{GSM8K} \\
        \midrule 
         Ground-Truth & 1 & 7473 & 121.8\\
        MixChain-C & 1 & 22419 & 294.8\\
        \rowcolor{gray!10} & 0 (Ground-Truth) & & 116.0\\
         \rowcolor{gray!10} & 1 &  & 279.6 \\
         \rowcolor{gray!10} & 2 &  & 310.7 \\
         \rowcolor{gray!10} & 3 &  & 386.7 \\
         \rowcolor{gray!10} \multirow{-5}{*}{MixChain-Z} & 4 &  \multirow{-5}{*}{6863}& 497.2 \\
        \midrule
        \multicolumn{4}{c}{PRM12K} \\
        \midrule
        Ground-Truth & 1 & 12000 & 223.1 \\
        \rowcolor{gray!10} & 0 (Ground-Truth) &  & 172.3\\
        \rowcolor{gray!10}& 1 &  & 583.2 \\
        \rowcolor{gray!10}& 2 &  & 613.7 \\
        \rowcolor{gray!10}& 3 &  & 739.3 \\
        \rowcolor{gray!10} \multirow{-5}{*}{MixChain-Z}& 4 &  \multirow{-5}{*}{8841} & 1003.2\\
        \midrule
        \multicolumn{4}{c}{LIMO} \\
        \midrule
        Ground-Truth & 1 & 817 & 6984.1 \\
        \rowcolor{gray!10} & 1 & 474 & 2994.7\\
        \rowcolor{gray!10} \multirow{-2}{*}{MixChain-C} & 2 & 564 & 4890.6 \\
        \bottomrule

    \end{tabular}
    }
    \caption{Dataset Statistic. Here we use the tokenizer from QwQ-32B-Preview to count the number of tokens.}
    \label{tbl:dataset}
    \end{center}
\end{table}

\section{More Analysis}

\paragraph{Experiments on DoRA.} In addition to LoRA, we also train LLaMA-3.2-1B using DoRA~\citep{DoRA} and control the magnitude of $\Delta \theta$ by adjusting the $\alpha$ for DoRA. The model is trained on QwQ synthesized data for a maximum of five epochs. We set the batch size to 8 and the peak learning rate to 4e-5, following a cosine decay schedule. A weight decay of 0.01 is applied. For DoRA, the rank is set to 32, and the lora\_alpha for training is set to 64. 

As shown in Table~\ref{tab:dora_gsm8k}, the chain length increases with the $\alpha$ value, demonstrating the effectiveness of interpolating $\Delta\theta$ for DoRA. Furthermore, similar to our observations with LoRA, the best result is not obtained by directly training the model on long CoT data. Specifically, training on QwQ synthesized data ($\alpha=1.0$) achieves an accuracy of 54.5 with 772.7 tokens, whereas the best model obtained via CoT-Valve ($\alpha$=0.5) achieves an accuracy of 55.72 with only 257.7 tokens.

\begin{table}[t]
    \centering
    \resizebox{\linewidth}{!}{
    \begin{tabular}{c|c|cccc|c}
    \toprule
        $\alpha$  & 0 & 0.125 & 0.25 &  0.5 & 0.75 & 1.0\\
        \midrule
        \# Tokens & 199.8 & 219.4 & 233.4 & 257.7 & 466.3 & 772.7\\
        Accuracy & 45.9 & 47.5 & 50.2 & 57.1 & 55.0 & 54.5 \\
        \bottomrule
    \end{tabular}
    }
    \caption{Results of LLaMA-3.2-1B-Instruct trained with DoRA using different $\alpha$ values for interpolation.}
    \label{tab:dora_gsm8k}
\end{table}

\begin{figure*}[t!]
\begin{tcolorbox}[parbox=false,colback=lightgreen!10, colframe=lightgreen!50, title=Generated Answer for the Same Question, coltitle=black]
\textbf{Question}: Amanda had 10 notebooks. This week, she ordered 6 more and then lost 2. How many notebooks does Amanda have now?

\rule{\linewidth}{0.4pt}

\noindent\textbf{Ground Truth Solution - 49 Tokens}:

This week, Amanda added 6 notebooks - 2 notebooks = <<6-2=4>>4 notebooks. Now, Amanda has 10 notebooks + 4 notebooks = <<10+4=14>>14 notebooks.

\rule{\linewidth}{0.4pt}

\noindent\textbf{Extrapolation - 300 Tokens}:
So Amanda started with 10 notebooks. Then, she ordered 6 more. So, I think I need to add those 6 to the original 10. That should be 10 plus 6, which is 16. Okay, now she lost 2 notebooks. So, I need to subtract those 2 from the 16. That should be 16 minus 2, which equals 14. So, Amanda has 14 notebooks now.
\textbf{\textit{\underline{Wait a minute, let me double-check that.}}} She had 10, ordered 6 more, so that's 10 + 6 = 16. Then lost 2, so 16 - 2 = 14. Yeah, that seems right.\textbf{\textit{\underline{Alternatively, maybe I can think of it differently}}}. Maybe consider the net gain of notebooks. She ordered 6 and lost 2, so net gain is 6 - 2 = 4. Then, add that net gain to the original 10: 10 + 4 = 14. Same answer. 
\textbf{\textit{\underline{Or, maybe another way:}}} starting with 10, losing 2 would bring her to 8, and then ordering 6 more would bring her to 14. Again, the same result.
I think no matter how I look at it, the answer is 14 notebooks.
**Final Answer**  \textbackslash boxed\{14\} 

\rule{\linewidth}{0.4pt}
\noindent\textbf{QwQ-32B-Preview - 257 tokens}:
So Amanda started with 10 notebooks. Then, she ordered 6 more. So, I need to add those 6 to the original 10. That should be 10 plus 6, which is 16. Okay, now she lost 2 notebooks. So, I need to subtract those 2 from the 16 she had after ordering more. That means 16 minus 2, which equals 14. So, Amanda has 14 notebooks now. \textbf{\textit{\underline{Wait a minute, let me double-check that}}}. She had 10, ordered 6 more, so that's 10 + 6 = 16. Then lost 2, so 16 - 2 = 14. Yeah, that seems right.   \textbf{\textit{\underline{Alternatively, maybe I can think of it differently}}}. Maybe combine the ordered and lost notebooks first. She ordered 6 and lost 2, so overall, she gained 6 - 2 = 4 notebooks. Then, add that to the original 10: 10 + 4 = 14. Same answer. I think that's correct. Amanda has 14 notebooks now. **Final Answer** \textbackslash boxed\{14\} 

\rule{\linewidth}{0.4pt}

\noindent\textbf{Interpolation - 207 Tokens}: Let's break this down step by step. Amanda started with 10 notebooks. Then, she ordered 6 more. So, I need to add those 6 to the original 10. That would be 10 plus 6, which equals 16 notebooks. Okay, so now she has 16 notebooks. But then, she lost 2. So, I need to subtract those 2 from the 16. That would be 16 minus 2, which equals 14 notebooks. Therefore, Amanda now has 14 notebooks. \textbf{\textit{\underline{Wait a minute, let me double-check that.}}} She started with 10, ordered 6 more, so that's 10 + 6 = 16. Then lost 2, so 16 - 2 = 14. Yeah, that seems right. I think that's the answer: Amanda has 14 notebooks now.

\rule{\linewidth}{0.4pt}

\noindent\textbf{Interpolation - 133 Tokens}:
Let's solve this step-by-step: 1. **Initial number of notebooks**: Amanda started with 10 notebooks. 2. **Notebooks ordered**: This week, she ordered 6 more notebooks.   - So, we add these 6 notebooks to her initial 10 notebooks.   - 10 + 6 = 16 notebooks. 3. **Notebooks lost**: Then, Amanda lost 2 notebooks.   - We need to subtract these 2 lost notebooks from the total she had after ordering more.   - 16 - 2 = 14 notebooks. Therefore, Amanda now has 14 notebooks.

\rule{\linewidth}{0.4pt}

\noindent\textbf{Extrapolation - 103 Tokens}: To find out how many notebooks Amanda has now, we can break the problem into steps:
1. Amanda initially had 10 notebooks.
2. This week, she ordered 6 more notebooks, so we add 6 to the initial amount: 10 + 6 = 16 notebooks.
3. Amanda then lost 2 notebooks, so we subtract 2 from the total: 16 - 2 = 14 notebooks.
So, Amanda has 14 notebooks now.

\end{tcolorbox} 
\caption{An example of the generated solution for the same question.} \label{fig:example}
\end{figure*}

\paragraph{Attention has less effect on the length of the reasoning path than MLP.}
We experimented with applying LoRA fine-tuning to only specific model components in QwQ. Table \ref{tbl:lora-position} shows that fine-tuning only the query, key, or value projection had less effect than other linear projections on shortening the reasoning chain.
This suggests that attention computation may have less impact on controlling chain length, whereas the MLP layers and the final projection in attention play a more significant role. 
We find this to be an intriguing phenomenon, and further investigation into its underlying causes could provide deeper insights into optimizing reasoning chain compression.

\begin{table}[t]
    \begin{center}
    \resizebox{\linewidth}{!}{
    \begin{tabular}{l|ccc|c}
        \toprule
        Modules & GSM8K & \#Tokens & \#Params & ACU $\uparrow$ \\
        \midrule 
        - & 95.1 & 741.1 & - & 0.40 \\
        \midrule
        K+V &  95.0 & 687.7 & 0.005\% & 0.43\\
        Q & 95.2 & 621.4 & 0.004\% & 0.48 \\
        O & 95.2 & 484.2 & 0.004\% & 0.61 \\
        Attention  & 94.2 & 284.2 & 0.013\% & 1.04 \\
        MLP        & 93.5 & 221.8 & 0.038\% & 1.32 \\
        \midrule
        All Linear & 92.4 & 227.6 & 0.051\% & 1.27 \\
        \bottomrule
    \end{tabular}
    }
    \caption{LoRA on Different Modules.}
    \label{tbl:lora-position}
    \end{center}
\end{table}

\paragraph{Prompt Control.}
We evaluate the length of CoT under constraint in prompts. Table \ref{tbl:prompt-control} presents the number of tokens generated when using various prompts across two models. For these two models, the prompts used here are:

\begin{tcolorbox}[parbox=false,colback=lightgreen!10, colframe=lightgreen!50, title=Prompt Template for QwQ-32B-Preview, coltitle=black]
\noindent\textbf{System}: You are a helpful and harmless assistant. You are Qwen developed by Alibaba. You should think step-by-step.

\noindent\textbf{User}: Generate the solution in less than <token\_count> tokens. <Question>
\end{tcolorbox}

\begin{tcolorbox}[parbox=false,colback=lightgreen!10, colframe=lightgreen!50, title=Prompt Template for LLaMA-3.2-1B-Instruct, coltitle=black]
\noindent\textbf{System}: 

\noindent Cutting Knowledge Date: December 2023

\noindent Today Date: 01 Jan 2025

\noindent\textbf{User}: 

\noindent Given the following problem, reason and give a final answer to the problem using less than <token\_count> tokens.

\noindent <Question>
\end{tcolorbox}

From the results, we observe that while these prompts provide control over the length, there remains a significant discrepancy between the generated token count and the intended target.

\section{Examples}
Here we show in Fig.\ref{fig:example} an example of the generated CoT from short to long, and we also show two extrapolation cases to show the generalization ability of our method. 
Our method notably generates a longer reasoning process compared to the original QwQ model, incorporating an extra reflection phase. During the chain shortening process, it reduces multiple rounds of reasoning and streamlines the language, ultimately enabling us to produce an answer with only 103 tokens through extrapolation.

\begin{table}[t!]
    \begin{center}
    \resizebox{\linewidth}{!}{
    \begin{tabular}{cc|cc}
        \toprule
        \multicolumn{2}{c|}{QwQ-32B-Preview} & \multicolumn{2}{c}{Llama-3.2-1B Instruct} \\
        \midrule
        Token in Prompt & \#Token Generated & Token in Prompt & \#Token Generated  \\
        \midrule 
        20 & 355 &  50 & 118  \\
        50 & 422 &  100 & 132 \\
        100 & 511 & 200 & 141 \\
        200 & 569 & 300 & 160 \\
        300 & 623 & 400 & 183 \\
        400 & 666 & 500 & 186 \\ 
        \bottomrule
    \end{tabular}
    }
    \caption{Significant discrepancies exist between the conditions specified in the prompt and the number of generated tokens on GSM8k.}
    \label{tbl:prompt-control}
    \end{center}
\end{table}

\end{document}